\documentclass[runningheads]{llncs}
\usepackage[T1]{fontenc}
\usepackage{amssymb}
\usepackage{amsmath} 
\usepackage{algorithm}
\usepackage{algpseudocode}
\usepackage[caption=false,font=footnotesize]{subfig}
\usepackage{booktabs}
\usepackage{graphicx}
\begin{document}
\title{Flow-based Policy With Distributional Reinforcement Learning in Trajectory Optimization}
\author{Ruijie Hao\inst{1}\orcidID{0009-0007-6038-2214} \and
Longfei Zhang\inst{1}\orcidID{0000-0001-8259-5148} \and
Yang Dai\inst{1}\orcidID{0009-0001-4813-8274}\and
Yang Ma\inst{2}\orcidID{0000-0001-7067-5423}\and
Xingxing Liang\inst{1}\orcidID{0000-0002-3594-2167}\textsuperscript{*} \and
Guangquan Cheng\inst{1}\orcidID{0009-0009-6953-7192}\textsuperscript{*}}
\authorrunning{Hao R.  et al.}
\institute{College of Systems Engineering, National University of Defense Technology, Changsha 410073, China \and
Aviation University of Air Force, Changchun 130000, China
}
\maketitle              % typeset the header of the contribution
\footnotetext[1]{Corresponding author.}

\begin{abstract}
Reinforcement Learning (RL) has proven highly effective in addressing complex control and decision-making tasks. However, in most traditional RL algorithms, the policy is typically parameterized as a diagonal Gaussian distribution, which constrains the policy from capturing multimodal distributions, making it difficult to cover the full range of optimal solutions in multi-solution problems, and the return is reduced to a mean value, losing its multimodal nature and thus providing insufficient guidance for policy updates. In response to these problems, we propose a RL algorithm termed flow-based policy with distributional RL (FP-DRL). This algorithm models the policy using flow matching, which offers both computational efficiency and the capacity to fit complex distributions. Additionally, it employs a distributional RL approach to model and optimize the entire return distribution, thereby more effectively guiding multimodal policy updates and improving agent performance. Experimental trails on MuJoCo benchmarks demonstrate that the FP-DRL algorithm achieves state-of-the-art (SOTA) performance in most MuJoCo control tasks while exhibiting superior representation capability of the flow policy. 
\end{abstract}
\section{Introduction}
\label{sec:introduction}
Recently, Reinforcement Learning (RL) has achieved remarkable success across a wide range of domains \cite{liu2024attention,zhualphaqcm}. In most existing RL algorithms, the policy is parameterized as a deterministic function or a Gaussian distribution and the return is regarded as the mean\cite{SAC}. However, the theoretically optimal policy may exhibit strong multimodality, which cannot be sufficient modeled by deterministic or Gaussian policies, and the return of state-action pairs also exhibit strong multimodality \cite{bellemare2017C51}, which is ignored by most existing algorithms. Restricted policy representations like unimodal Gaussians can trap algorithms in local optimal by exhibiting mode-covering behavior that concentrates probability density in low value regions between distinct high value actions, thereby severely impairing policy performance. Additionally, if one policy cannot cover the full range of optimal solutions in current state, which may also not find the optimal solution in next state. Moreover, two returns with the same mean can have vastly different distributions. Conditioning policy learning solely on the mean thus discards critical information about return distribution, which is essential for distinguishing between actions that are consistently rewarding and those that are merely average in expectation. Consequently, modeling the policy with a unimodal Gaussian distribution and guiding it by the mean of returns is likely to significantly impair policy learning. 

Lately, the generative model has been widely used in policy modeling for its powerful ability to fit multimodal distributions \cite{rombach2022high,geng2025mean}. In offline RL, expert behaviors can exhibit significant variability even in the same state. This necessitates the use of generative models capable of fitting multimodal distributions\cite{janner2022planning}. In Online RL, policy also need to be multimodal to fit all theoretically optimal actions. This allows the agent to achieve superior performance by capturing diverse optimal behaviors and adapting more flexibly to dynamic environments\cite{psenka2024learning}. However, most existing algorithms employ diffusion models as policies, which require numerous steps to generate the target distribution, making it difficult to meet the real-time constraints of complex control tasks. Additionally, the majority of existing algorithms treat the return as a mean value, failing to provide sufficient guidance for policies with complex distributions. Therefore, this paper focuses on how the Policy can be trained as Flow-based policy, which requires only few steps inference to generate actions, and investigates the performance of guiding such policies via distributional RL.

In this work, we propose flow-based policy with distributional RL (FP-DRL), a generalized new approach to combine flow-based policy with distributional critic in online RL. A recent work by Wang et al. \cite{wang2024diffusion} only uses the reverse diffusion processes to approximate policy and the optimization objective is to maximize the expected Q-value, which ignores the loss derived from supervised learning of diffusion model. Inspired by this work, We utilize only the network architecture of the flow model to enable the model to generate multimodal distributions, and train it directly using RL objectives, without incorporating a loss function between current actions and target actions. Then, we integrate distributional RL and employ the distributional critic to guide the policy update more effectively.

In summary, the key contributions of this paper are the following: 1) We propose to  approximate the policy using only the architecture of the Flow Matching (FM) model, training it directly with RL objectives rather than with the objectives derived from supervised learning. 2) We introduce distributional RL and leverage the representation of the full return distribution to provide more discriminative guidance for the multimodal policy. 3) We evaluate the efficiency, generality and performance of our method on the popular MuJoCo \cite{todorov2012mujoco} benchmarking. Compared with SAC-Flow \cite{zhang2025sac}, DSAC-T\cite{duan2025distributional}, DACER\cite{wang2024diffusion}, SAC\cite{SAC}, TD3\cite{TD3}, our approach achieves the SOTA performance. 4) We provide the FP-DRL code written in JAX and Pytorch to facilitate future researchers to follow our work.

Section \ref{sec:relatedworks} reviews generative policies and distributional RL, motivating our approach. Section \ref{sec:preliminaries} outlines the theoretical foundations, including flow matching and distributional RL. Section \ref{sec:method} details the FP-DRL framework, integrating flow-based policies with a distributional critic in Soft Actor-Critic. Section \ref{sec:experiments} evaluates FP-DRL on MuJoCo benchmarks against baselines and via ablation studies. Finally, Section \ref{sec:conclusion} summarizes our contributions and findings.

\section{Related Works}
\label{sec:relatedworks}
\subsection{Generative Policy in RL}
\subsubsection{Diffusion Policy in RL} Diffusion models were first employed in offline RL for cloning multiple expert behaviors. Cheng et al. \cite{chi2025diffusion} proposed a visuomotor policy based on conditional denoising diffusion, which achieves superior performance in robotic manipulation tasks by effectively modeling multimodal action distributions and enabling stable training. Wang et al. \cite{wangdiffusion} introduced Diffusion-QL, which uses a diffusion model as an expressive policy class and integrates Q-learning guidance to achieve effective policy regularization and improvement. The original diffusion models need labels to train, while expert behaviors can serve as labels in offline RL, they are not available in online RL. Wang et al. \cite{wang2024diffusion} proposed the DACER algorithm, which leverages the reverse process of the diffusion model as the policy function, estimates entropy using a Gaussian mixture model, and directly employs Q-values to guide policy updates without requiring labels, making it applicable to online RL.

\subsubsection{Flow Policy in RL} Although diffusion policy can fit complex multimodal distributions, it suffers from low efficiency due to numerous steps sampling, making it inadequate for real-time control tasks. A flow model learns a velocity field, enabling sampling from an initial simple distribution to a complex multimodal distribution in just a few steps, or even a single step \cite{zhangreinflow,celik2025dime}. Research on flow-based policies is just beginning to emerge. Jiang et al. \cite{jiangstreaming} proposed Streaming Flow Policy, which treats action trajectories as flow trajectories and enables streaming generation from a narrow Gaussian initialization while preserving multimodality. Zhang et al. \cite{zhang2025sac} proposed SAC-flow, which stabilizes off-policy training of flow-based policies by reparameterizing the velocity network as a GRU or Transformer to mitigate gradient pathologies.
\subsection{Distributional RL}
\subsubsection{Distributional RL In value-based RL} Distributional RL captures the full distribution of returns rather than just the expected value, providing a richer learning signal that enables more stable, efficient, and risk-aware policy optimization \cite{bellemare2023distributional,muller2024distributional,lowet2025opponent}. The groundbreaking paper of distributional RL is C51 \cite{bellemare2017C51} algorithm proposed by Bellmare et al. which parametrizes the return distribution as a categorical distribution over a discrete set of fixed supports, learning the probability mass per atom through minimization of the KL divergence. Quantile-based methods instead learn where the supports of the distribution should lie. QRDQN \cite{dabney2018distributional} uses a uniform mixture of Dirac delta functions to approximate the return distribution, where each quantile level is assigned equal probability. This formulation allows the minimization of the quantile regression loss, which is an approximation of the 1-Wasserstein distance. Further extensions, such as Implicit Quantile Networks (IQN) \cite{dabney2018implicit} or Fully Quantile Functions (FQF) \cite{FQF} enhance the expressiveness of the learned distribution. IQN samples quantile fractions from some underlying base distribution and FQF learns the location of the fractions in an end-to-end manner.
\subsubsection{Distributional RL With Actor-Critic Integration} Recent works have extended distributional RL to continuous control problems using the Actor-Critic (AC) framework \cite{xiao_MDSAC_2024,wiltzer2024foundations,wiltzer2024action}. Prior work, such as D4PG \cite{barth-maron2018D4PG}, adapted the categorical projection from C51, but inherited the limitations of fixed supports and the use of KL divergence. More recently, methods like DSAC \cite{duan2021distributional} and DSAC-T~\cite{duan2025distributional} integrated the AC framework by modeling returns with a Gaussian distribution. However, a critical limitation persists: these methods optimize via Maximum Likelihood Estimation (MLE), which effectively minimizes the KL divergence rather than the contractive Wasserstein metric. 

Our method is motivated to propose a flow-based policy and a distributional critic that can be combined with most existing actor-critic frameworks. We first use the flow model as a policy function with strong representational power. Then, we construct a distributional critic to provide sufficient guidance for the multimodal distribution policy and improve the performance of the policy.

\section{Preliminaries}
\label{sec:preliminaries}
\subsection{Online RL}
In online RL, the agent interacts with environment in real time, and this interaction can be modeled as a Markov Decision Process (MDP) defined by the tuple $(\mathcal{S,A},R,P,\gamma)$ \cite{puterman1994markov,wang2022model}. Here, $\mathcal{S}$ denotes the state space, $\mathcal{A}$ denotes the action space, R denotes the reward function under the state and action, $P$ denotes the transition probability from state $\mathcal{S}$ to $\mathcal{S}'$ under action $a$. $\gamma \in (0,1)$ denotes the reward discount factor.

For an agent following a fixed policy $\pi$, the discounted return is denoted by the random variable:
\begin{equation}
Z^{\pi}(s,a) = \sum_{t=0}^{\infty} \gamma^t R(s_t, a_t),
\end{equation}
where $s_0 = s$, $a_0 = a$, $s_t \sim P(\cdot|s_{t-1}, a_{t-1})$ and $a_t \sim \pi(\cdot|s_t)$.
The action-value function is defined as $Q^{\pi}(s,a) = \mathbb{E}\left[Z^{\pi}(s,a)\right]$ and can be characterized by the Bellman equation:
\begin{equation}
Q^{\pi}(s,a) = \mathbb{E}\left[R(s,a)\right] + \gamma \mathbb{E}_{P,\pi}\left[Q^{\pi}(s',a')\right].
\end{equation}

The objective in RL is to find an optimal policy $\pi^*$, which maximizes $\mathbb{E}[Z^{\pi}]$ and satisfies $Q^{\pi^*}(s,a) \geq Q^{\pi}(s,a)$ for all $\pi$ and all $s,a$. 
\subsection{Maximum Entropy RL}
Standard RL aims to maximize the expected cumulative reward \cite{schulman2017PPO}. We consider a more general maximum entropy objective that augments the original objective with the expected entropy of the policy \cite{haarnoja2017reinforcement}:
\begin{equation}
    J_\pi = \mathbb{E}_{(s_{i \geq t}, a_{i \geq t}) \sim \rho_\pi} \left[ \sum_{i=t}^{\infty} \gamma^{i-t} [r_i + \alpha \mathcal{H}(\pi(\cdot \mid s_i))] \right],
    \label{eq:entropy_objective}
\end{equation}
where $\gamma \in (0,1)$ is the discount factor, $\alpha$ is the temperature coefficient and the policy entropy $\mathcal{H}$ is expressed as
\[
    \mathcal{H}(\pi(\cdot \mid s)) = \mathbb{E}_{a \sim \pi(\cdot \mid s)} [-\log \pi(a \mid s)].
\]

We denote the entropy-augmented accumulated return from $s_t$ as $G_t = \sum_{i=t}^\infty \gamma^{i-t} [r_i - \alpha \log \pi(a_i \mid s_i)]$, also referred to as soft return. The soft Q of $\pi$ is given as
\begin{equation}
    Q^\pi(s_t, a_t) = r_t + \gamma \mathbb{E}_{(s_{i>t}, a_{i>t}) \sim \rho_\pi} [G_{t+1}],
    \label{eq:soft_q}
\end{equation}
which defines the expected soft return for choosing $a_t$ in state $s_t$ and subsequently following policy $\pi$.

We solve for the optimal policy using soft policy iteration, a method consisting of two interacting processes: evaluation and improvement. Soft policy evaluation fits the soft Q-function to the current policy $\pi$ by iterating the Bellman operator $\mathcal{T}^\pi$:
\begin{equation}
\mathcal{T}^\pi Q^\pi(s,a) = r + \gamma \, \mathbb{E}_{s' \sim p,\, a' \sim \pi} \big[  Q^\pi(s', a')  - \alpha \log \pi(a' \mid s') \big].
\end{equation}

Soft policy improvement updates the policy by maximizing the expected soft Q-value. The updated policy $\pi{\text{new}}$ is obtained via:
\begin{equation}
\pi_{\text{new}} = \arg\max_\pi \, \mathbb{E}_{s \sim \rho_\pi,\, a \sim \pi} \big[  Q^{\pi_{\text{old}}}(s,a) - \alpha \log \pi(a \mid s) \big].
\label{eq:policy_update}
\end{equation}

As established in prior work \cite{haarnoja2017reinforcement}, alternating between these evaluation and improvement steps ensures convergence to the global optimum.

\subsection{Distributional RL}
Unlike traditional value-based RL, which estimates the scalar expectation of the cumulative return $Q^\pi(s,a) = \mathbb{E}[Z^\pi(s,a)]$, distributional RL seeks to model the full probability distribution of the random return $Z^\pi(s,a)$.

The distribution of returns satisfies the distributional Bellman equation:
\begin{equation}
    Z^\pi(s,a) \overset{D}{=} R(s,a) + \gamma Z^\pi(S',A'),
    \label{eq:dist_bellman}
\end{equation}
where $S' \sim P(\cdot|s,a)$, $A' \sim \pi(\cdot|S')$ and $\overset{D}{=}$ denotes equality in distribution. The corresponding distributional Bellman operator $\mathcal{T}^\pi$ is defined as
\begin{equation}
    \mathcal{T}^\pi Z(s,a) {:=} R(s,a) + \gamma Z(S',A').
\end{equation}

For the control setting, the optimal return distribution $Z^*$ is the fixed point of the distributional optimality operator $\mathcal{T}$:
\begin{equation}
    \mathcal{T}Z(s,a) {:=} R(s,a) + \gamma Z(S', \pi^*(S')),
\end{equation}
where $\pi^*$ is the greedy policy that maximizes the expected return, i.e., $\pi^*(s') = \arg\max_{a' \in \mathcal{A}} \mathbb{E}[Z(s', a')]$.
\subsection{Flow Matching}
\begin{figure}
\centering
\includegraphics[width=0.75\textwidth]{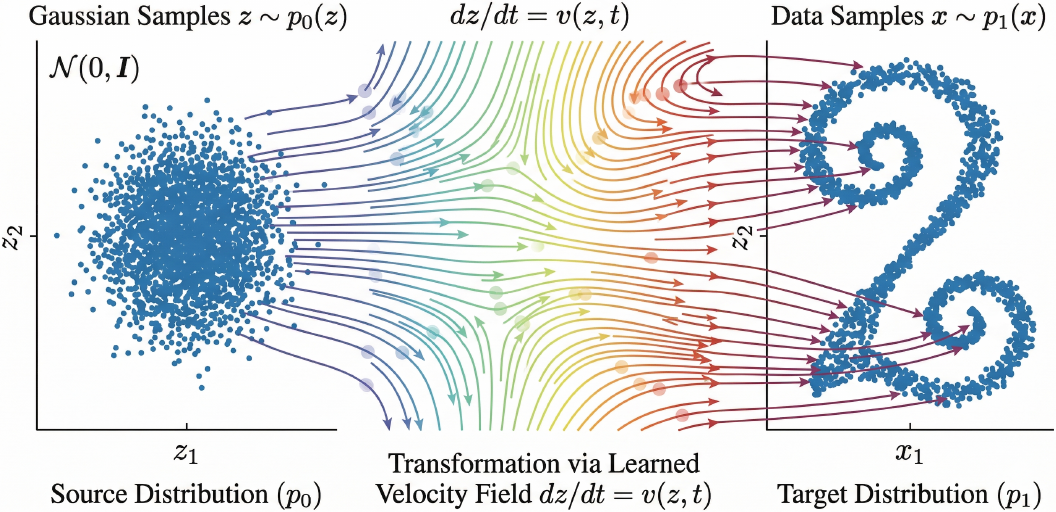}
\caption{FM learns a velocity fields that transports samples from a simple prior distribution to the target data distribution via an ordinary differential equation (ODE).} \label{fig1}
\end{figure}

FM is a family of generative models that learn to match the velocity fields between two probabilistic distributions, where the velocity is regarded as the flow \cite{papamakarios2021normalizing,liu2023flow}. Unlike diffusion models \cite{ho2020denoising}, which require a two step process of first adding noise and then denoising, FM can directly transition from an initial simple distribution, such as a Gaussian, to a complex target distribution. The process of FM is shown as Fig.~\ref{fig1}.

Given data $x \sim p_{\text{1}}(x)$ and prior $\epsilon \sim p_{\text{0}}(\epsilon)$, a flow path can be constructed as $z_t = a_t x + b_t \epsilon$ with time $t$, where $a_t$ and $b_t$ are predefined schedules. The velocity $v_t$ is defined as $v_t = z'_t = a'_t x + b'_t \epsilon$, where $'$ denotes the time derivative. The schedule used in Rectified Flow and now widely adopted is $a_t = 1 - t$ and $b_t = t$, which leads to $v_t = \epsilon - x$. 

Because a given $z_t$ and its $v_t$ can arise from different $x$ and $\epsilon$, FM essentially models the expectation over all possibilities, called the marginal velocity:
\begin{equation}
v(z_t, t) \triangleq \mathbb{E}_{p_t(v_t|z_t)}[v_t].
\end{equation}

A neural network $v_\theta$ parameterized by $\theta$ is learned to fit the marginal velocity field: $\mathcal{L}_{\text{FM}}(\theta) = \mathbb{E}_{t,p_t(z_t)} \| v_\theta(z_t,t) - v(z_t,t) \|^2$. Although computing this loss function is infeasible due to the marginalization in Eq.~(10), it is proposed to instead evaluate the conditional FM loss [28]: $\mathcal{L}_{\text{CFM}}(\theta) = \mathbb{E}_{t,x,\epsilon} \| v_\theta(z_t,t) - v_t(z_t \mid x) \|^2$, where the target $v_t$ is the conditional velocity. Minimizing $\mathcal{L}_{\text{CFM}}$ is equivalent to minimizing $\mathcal{L}_{\text{FM}}$ [28].

Given a marginal velocity field $v(z_t, t)$, samples are generated by solving an ODE for $z_t$:
\begin{equation}
\frac{d}{dt} z_t = v(z_t, t)
\end{equation}
starting from $z_0 = \epsilon \sim p_{\text{0}}$. The solution can be written as: $z_r = z_t - \int_r^t v(z_\tau, \tau) d\tau$, where we use $r$ to denote another time step. In practice, this integral is approximated numerically over discrete time steps. For example, the Euler method, a first-order ODE solver, computes each step as: $z_{t_{i+1}} = z_{t_i} + (t_{i+1} - t_i) v(z_{t_i}, t_i)$. Higher-order solvers can also be applied.

\section{Method}
\label{sec:method}
In this section, we detail the design of our flow-based policy with distributional RL. First, we use FM as policy approximator, serving as the policy function in RL. Second, we directly optimize the flow-based policy using gradient descent, whose objective function is to maximize expected Q-values. At this point, the return is still regarded as the mean. However, the mean loses critical information such as risk and multimodality; only the full return distribution can provide all the necessary guidance for risk-aware, robustly optimized, and stably learned policies. Consequently, we employ a quantile-based approach to model the full return distribution and utilize distributional RL to update the returns.

\subsection{Flow-based Policy Representation}
We conceptualize the flow-based policy as a conditional FM conditioned on states S. Starting from an initial Gaussian distribution $p_0(\cdot \mid s) \sim \mathcal{N}(0,I)$ on an action space $\mathcal{A} = \mathbb{R}^d$, the target policy $p_1(\cdot \mid s)$ is reached via a time-indexed mapping $\varrho : [0, 1] \times \mathcal{A} \times \mathcal{S} \to \mathcal{A}$, with $A_t := \varrho_t(A_0 \mid s)$ for $t \in [0, 1]$, where $A_0 \sim p_0(\cdot \mid s)$ and $A_1 \sim p_1(\cdot \mid s)$. 

The trajectory satisfies the ODE $\frac{\mathrm{d}}{\mathrm{d}t}\varrho_t(A_0 \mid s) = v(t, \varrho_t(A_0 \mid s), s)$, where $v$ is the velocity field to be learned. The learned field is integrated numerically with flow rollout to obtain:
\begin{equation}
A_{t_{i+1}} = A_{t_i} + \Delta t_i \, v_\theta(t_i, A_{t_i}, s), \quad 0 = t_0 < \cdots < t_K = 1,
\end{equation}

Where $\Delta t_i = t_{i+1} - t_i$. The resulting distribution over $A_1$ induced by $A_0 \sim \mathcal{N}(0, I_d)$ is denoted $\mu_\theta(\cdot \mid s)$ and serves as the stochastic policy $a = A_1 \sim \pi_\theta(\cdot \mid s)$.

We use Transformer to approximate the velocity field $v_\theta(t_i,A_{t_i},s)$, the model computes embeddings for the current action token $A_{t_i}$, time token $t_i$ and global embedding s.

\subsection{Distributional Critic Learning via Quantile Regression}
To provide more comprehensive guidance for the flow-based policy, we model the distributions of return rather than expected value which used by most existing methods. We models the distribution over returns by assigning fixed and uniform probabilities to adjustabl locations. We use a neural network to learn the locations $\delta_{\theta_{i(x,a)}}$. Then the quantile distribution $Z_\theta$ of each state-action pair $(x,a)$ can be represented as:
\begin{equation}
    Z_{\theta}(x, a) := \frac{1}{N} \sum_{i=1}^{N} \delta_{\theta_i(x, a)},
\end{equation}
where $\delta_z$ denotes a Dirac at $z \in R$.

\begin{lemma}[Bellemare et al., 2017]
\label{lemma:contraction}
The distributional Bellman operator $\mathcal{T}^\pi$ for policy evaluation is a $\gamma$-contraction under the maximal $p$-Wasserstein metric $\bar{d}_p$:
\begin{equation}
    \bar{d}_p(\mathcal{T}^\pi Z_1, \mathcal{T}^\pi Z_2) \leq \gamma \bar{d}_p(Z_1, Z_2).
\end{equation}
\end{lemma}

Lemma 1 shows that $\bar{d}_p$ is a rigorous metric for analyzing the convergence of distributional RL algorithms toward the fixed point $Z^\pi$. Based on Lemma 1, we can learn value distributions by minimizing the Wasserstein distance between $Z$ and its Bellman update $\mathcal{T}^\pi Z$, which is analogous to minimizing the $L^2$ distance in standard temporal difference (TD) learning to align $Q$ with $\mathcal{T}Q$.

We use huber loss to minimize the 1-Wasserstein distance of current return Z and target return $\mathcal{T}^\pi Z$:
\begin{equation}
\mathcal{L}(\theta) = \frac{1}{N} \sum_{i=1}^{N} \sum_{j=1}^{N'} \rho_{\tau_i}^\kappa \left( y_j - \theta_i(x, a) \right),
\end{equation}
where $y_j = r + \gamma \theta_j(x', a')$ represents the sampled target returns from $\mathcal{T}^\pi Z$, and the quantile Huber loss $\rho_\tau^\kappa(\delta)$ is defined as:
\begin{equation}\rho_\tau^\kappa(\delta) = \left| \tau - \mathbb{I}{\delta < 0} \right| \mathcal{L}_\kappa(\delta).\end{equation}

Here, $\mathbb{I}\{\cdot\}$ is the indicator function, and $\mathcal{L}_\kappa(\delta)$ is the standard Huber loss with threshold $\kappa$:
\begin{equation}
\mathcal{L}_\kappa(\delta) = \begin{cases}\frac{1}{2} \delta^2 & \text{if } |\delta| \leq \kappa, \\ \kappa \left( |\delta| - \frac{1}{2} \kappa \right) & \text{otherwise.}\end{cases}
\end{equation}

\subsection{Training via SAC}
Once the policy parameterized by FM and the distributional critic are established, we integrate them into the Soft Actor-Critic (SAC) framework. In SAC, the actor and critic are trained alternately to optimize the maximum entropy objective.

\subsubsection{Distributional Soft Critic Update}
In our FP-DRL framework, the critic models the full return distribution. To account for the maximum entropy objective, we augment the target return distribution with the policy entropy. For a sampled transition $(s, a, r, s')$, the target quantiles $y_j$ are computed as:
$$y_j = r + \gamma \big( \theta'_j(s', a') - \alpha \log \pi_\phi(a' \mid s') \big),$$
where $a' \sim \pi_\phi(\cdot \mid s')$, $\alpha$ is the temperature parameter, and $\theta'$ denotes the parameters of the target distributional critic network. The critic is updated by minimizing the quantile Huber loss between the current predicted quantiles and the target quantiles, as defined in Eq. (15). 

\subsubsection{Flow-based policy update}
We consider the maximum entropy objective and directly employ the RL objective for policy updates. Thus, the policy objective function is formulated as:
$$J_\pi(\phi) = \mathbb{E}_{s \sim \mathcal{D}, a \sim \pi_\phi} \left[ \alpha \log \pi_\phi(a \mid s) - \frac{1}{N} \sum_{i=1}^{N} \theta_i(s, a) \right],$$
where $\mathcal{D}$ is the replay buffer and $\phi$ represents the parameters of the flow-based policy.

To compute \(\log \pi_\phi(a | s)\) when the velocity field is only available at discrete times \(t_0,\dots,t_N\) with \(t_0=0,t_N=1\), we treat it as piecewise constant on each interval \([t_{i-1},t_i]\). Given a base Gaussian \(p_0\) and learned fields \(v_{\phi,i}(a,s)\), the change in log probability follows
\[
\frac{\partial \log p_t(a_t | s)}{\partial t} = -\operatorname{Tr}\bigl(\nabla_{a_t} v_{\phi,i}(t, a_t, s)\bigr), \quad t\in[t_{i-1},t_i].
\]

Integrating over all segments yields
\[
\log \pi_\phi(a_1 | s) = \log p_0(a_0 | s) - \sum_{i=1}^{N} \int_{t_{i-1}}^{t_i} \operatorname{Tr}\bigl(\nabla_{a_t} v_{\phi,i}(t, a_t, s)\bigr) \, dt,
\]
where \(a_t\) evolves as \(\frac{da_t}{dt}=v_{\phi,i}(t,a_t,s)\) on each segment. In practice, the state and log probability are integrated numerically. The trace can be computed exactly for low-dimensional actions or approximated via Hutchinson’s estimator for high-dimensional control tasks.

The adjustment of the temperature parameter $\alpha$ follows the standard SAC algorithm, where the parameter is automatically tuned to match a target entropy $\bar{\mathcal{H}}$, typically set to the negative dimension of the action space. The objective for $\alpha$ is:
$$J(\alpha) = \mathbb{E}_{s \sim \mathcal{D}, a \sim \pi_\phi} \left[ -\alpha \big( \log \pi_\phi(a \mid s) + \bar{\mathcal{H}} \big) \right].$$

By alternating the optimization of the distributional critic, the flow-based actor, and the temperature parameter, FP-DRL robustly learns multimodal policies guided by rich, distributional value signals. The complete procedures are summarized in Algorithm 1.

\begin{algorithm}[htbp]
\caption{FP-DRL Training via SAC}
\label{alg:fp_drl}
\begin{algorithmic}[1]
\State Initialize distributional critic $\theta$, target $\theta'$, flow-based policy $\pi_\phi$; replay buffer $\mathcal{D}$.
\For{each update}
    \State Interact with the environment using $\pi_\phi$; push $(s_t, a_t, r_t, s_{t+1})$ to $\mathcal{D}$.
    \State Sample $\{(s_h, a_h, r_h, s_{h+1})\}_{h=1}^B \sim \mathcal{D}$.
    \State Critic: compute target quantiles $y_j$ and minimize the quantile Huber loss (Equ. 15).
    \State Actor: draw $a \sim \pi_\phi$ and compute $\log \pi_\phi(a \mid s)$ via numerical integration of velocity field $v_{\phi,i}$; minimize $J_\pi(\phi)$.
    \State Temperature: automatically adjust $\alpha$ to match target entropy $\bar{\mathcal{H}}$ by minimizing $J(\alpha)$.
    \State Update target network $\theta'$ by an exponential moving average.
\EndFor
\end{algorithmic}
\end{algorithm}

\section{Experiments}
\label{sec:experiments}
We evaluate the performance of our method in locomotion and manipulation tasks of RL within MuJoCo. We conducted experiments on the most challenging benchmark tasks, which include Humanoid-v4, Ant-v4, HalfCheetah-v4, Hopper-v4, Reacher-v4 and InvertedPendulum-v4. 

\begin{figure*}[htbp]
\centering
\captionsetup[subfigure]{justification=centering}
\subfloat[Humanoid-v4\label{subFig:humanoid}]
{\includegraphics[width = 0.3\textwidth,height = 0.25\textwidth]{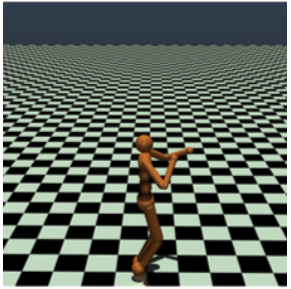}} \quad
\subfloat[Ant-v4\label{subFig:ant}]
{\includegraphics[width = 0.3\textwidth,height = 0.25\textwidth]{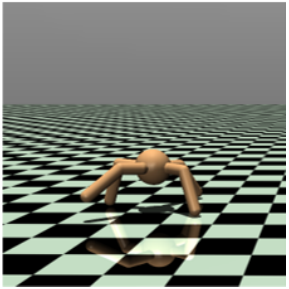}} \quad
\subfloat[Hopper-v4\label{subFig:halfcheetah}]
{\includegraphics[width = 0.3\textwidth,height = 0.25\textwidth]{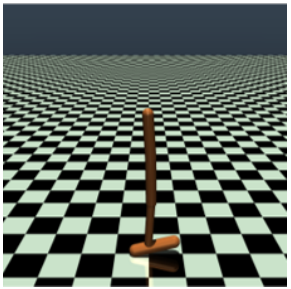}} \quad
\vspace{0.5em}
\subfloat[HalfCheetah-v4\label{subFig:humanoid}]
{\includegraphics[width = 0.3\textwidth,height = 0.25\textwidth]{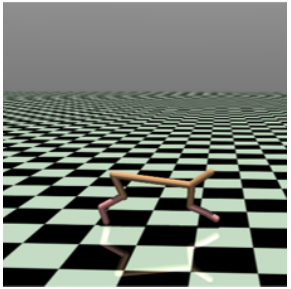}} \quad
\subfloat[InvertedPendulum-v4\label{subFig:ant}]
{\includegraphics[width = 0.3\textwidth,height = 0.25\textwidth]{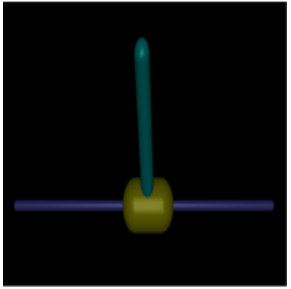}} \quad
\subfloat[Reacher-v4\label{subFig:Reacher}]
{\includegraphics[width = 0.3\textwidth,height = 0.25\textwidth]{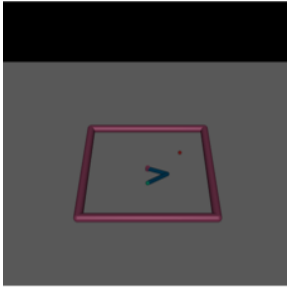}} \quad
\caption{Benchmarks. 
    (a) Humanoid-v4: \((s \times a) \in \mathbb{R}^{376} \times \mathbb{R}^{17}\). 
    (b) Ant-v4: \((s \times a) \in \mathbb{R}^{111} \times \mathbb{R}^{8}\).
    (c) Hopper-v4: \((s \times a) \in \mathbb{R}^{11} \times \mathbb{R}^{3}\).
    (d) HalfCheetah-v4: \((s \times a) \in \mathbb{R}^{17} \times \mathbb{R}^{6}\).
    (e) InvertedPendulum-v: \((s \times a) \in \mathbb{R}^{4} \times \mathbb{R}^{1}\).
    (f) Reacher-v4: \((s \times a) \in \mathbb{R}^{11} \times \mathbb{R}^{2}\).
}
\label{f:benchmeark}
\end{figure*}

\noindent\textbf{Benchmarks.} We measure its performance on 6 original, unmodified continuous control tasks from MuJoCo, interfaced through OpenAI Gym. Fig.~\ref{f:benchmeark} illustrates the benchmark tasks used in this study. 
% and more description of each task is provided in Appendix B.

\noindent\textbf{Baselines.} Our algorithm is compared and evaluated against the six well-known model-free algorithms. These include TD3, SAC, DSAC-T and SAC-Flow. These baselines have been extensively tested and applied in a series of demanding domains and prior to the arXiv submission of this article, SAC-Flow was the SOTA algorithm identified in our survey.

\noindent\textbf{Experimental Details.} To ensure a fair comparison, we reproduced all baseline algorithms in both JAX and PyTorch frameworks, achieving a 2–3× speedup in training compared to using pure PyTorch while maintaining identical performance. All algorithms and tasks employed the same three-layer MLP neural network with ReLU activation functions, and all parameter updates were performed using the Adam optimizer. The total number of training steps was set to one million for all experiments, and the reported results were averaged over three random seeds. The experiments were conducted on an Intel(R) Xeon(R) Gold 6430 32-core CPU and an NVIDIA GeForce RTX 4090 GPU. 
%Additional hyperparameter details are provided in Appendix A.2 due to space limitations.

\noindent\textbf{Evaluation Protocol}. This study adopts the same evaluation metrics as DACER and DSAC-T: for each random seed, the highest return observed within the last 10\% of training steps for each run is averaged, and each evaluation result is obtained by averaging over ten test episodes. The final results are aggregated across three seeds to compute the mean and standard deviation. Additionally, the training curves in Figure 1 illustrate the stability characteristics of the training process.

\subsection{Comparative Performance}

\begin{figure*}[htbp]
\centering
\captionsetup[subfigure]{justification=centering}
\subfloat[Ant-v4\label{subFig:humanoid}]
{\includegraphics[width = 0.3\textwidth]{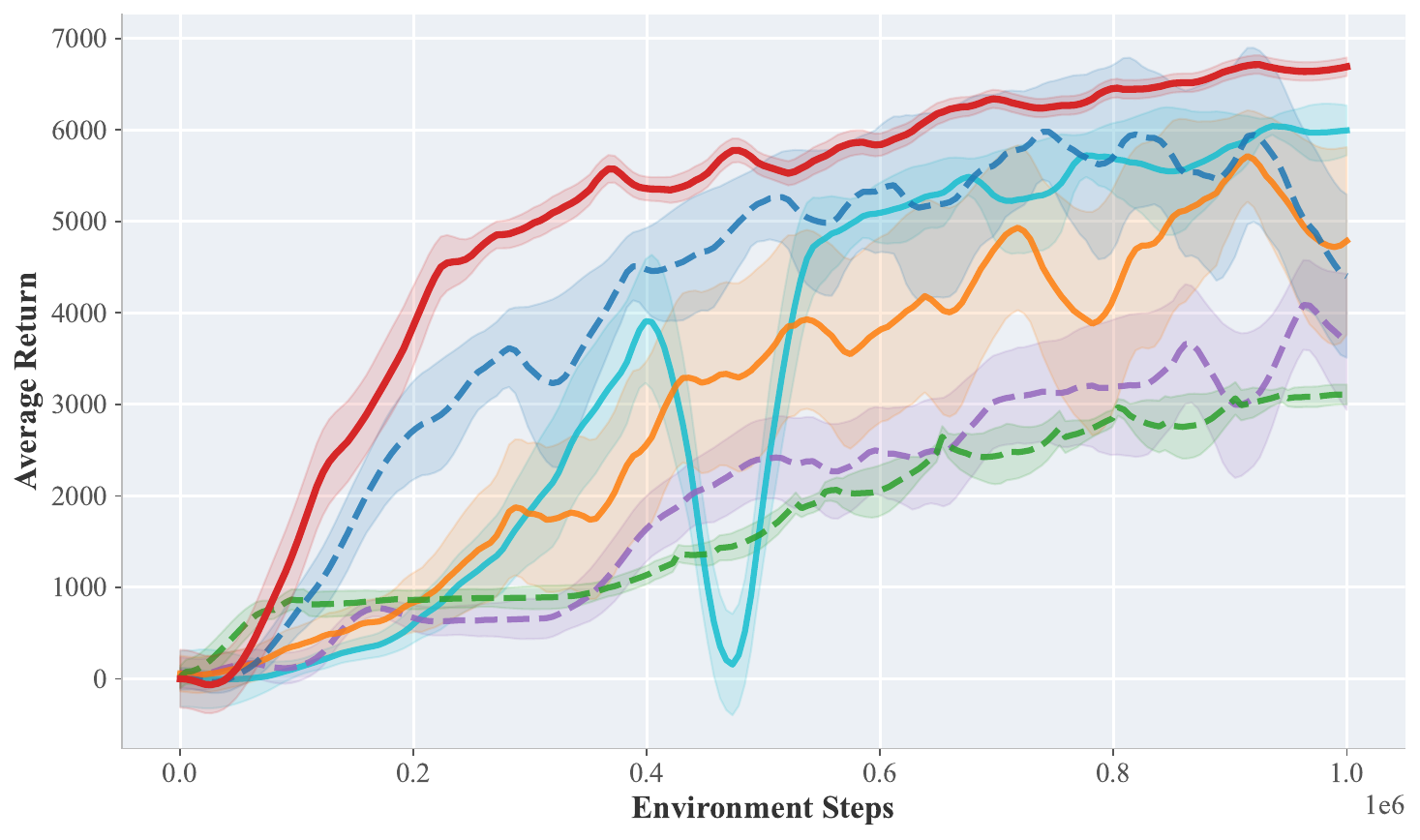}} \quad
\subfloat[HalfCheetah-v4\label{subFig:ant}]
{\includegraphics[width = 0.3\textwidth]{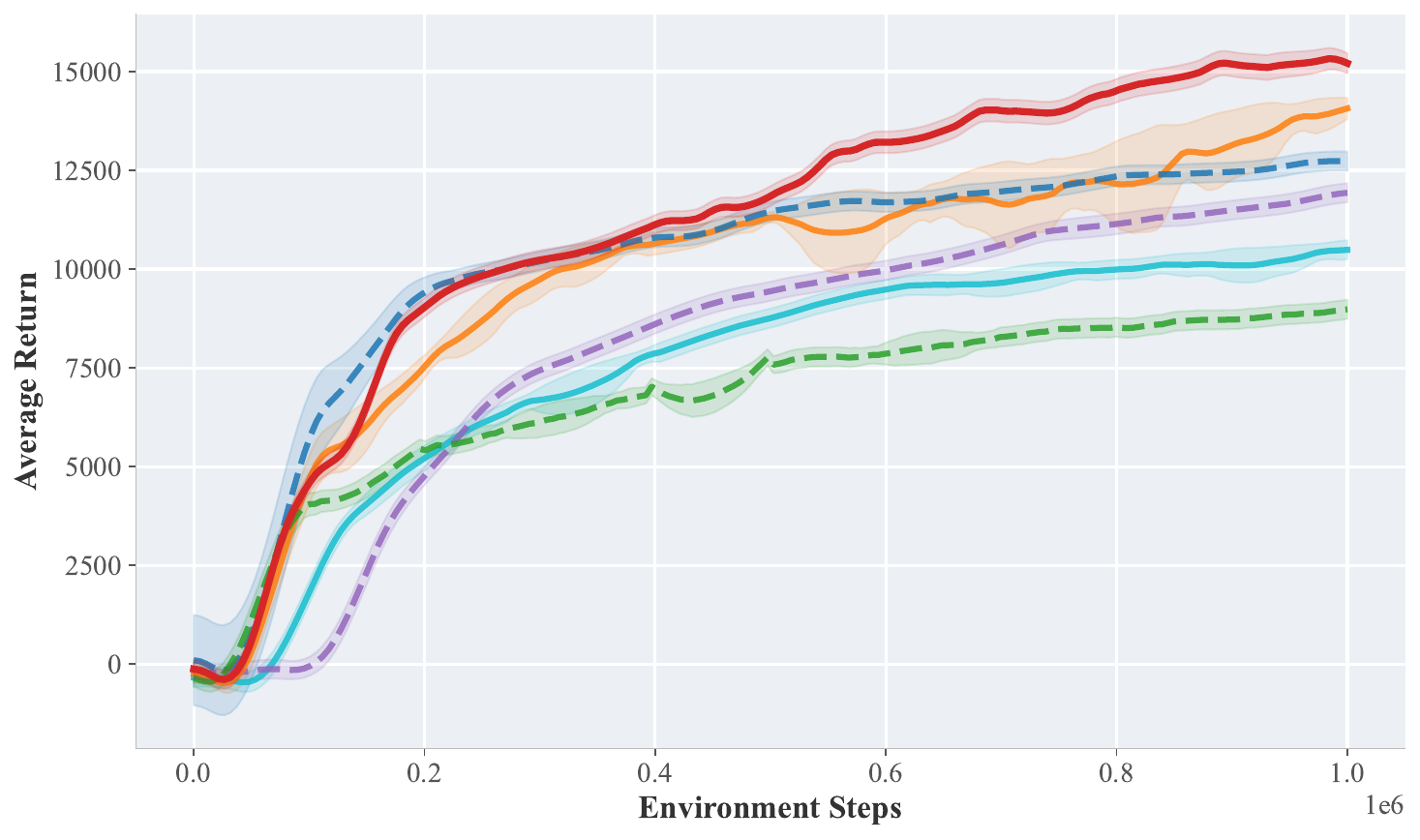}} \quad
\subfloat[Hopper-v4\label{subFig:halfcheetah}]
{\includegraphics[width = 0.3\textwidth]{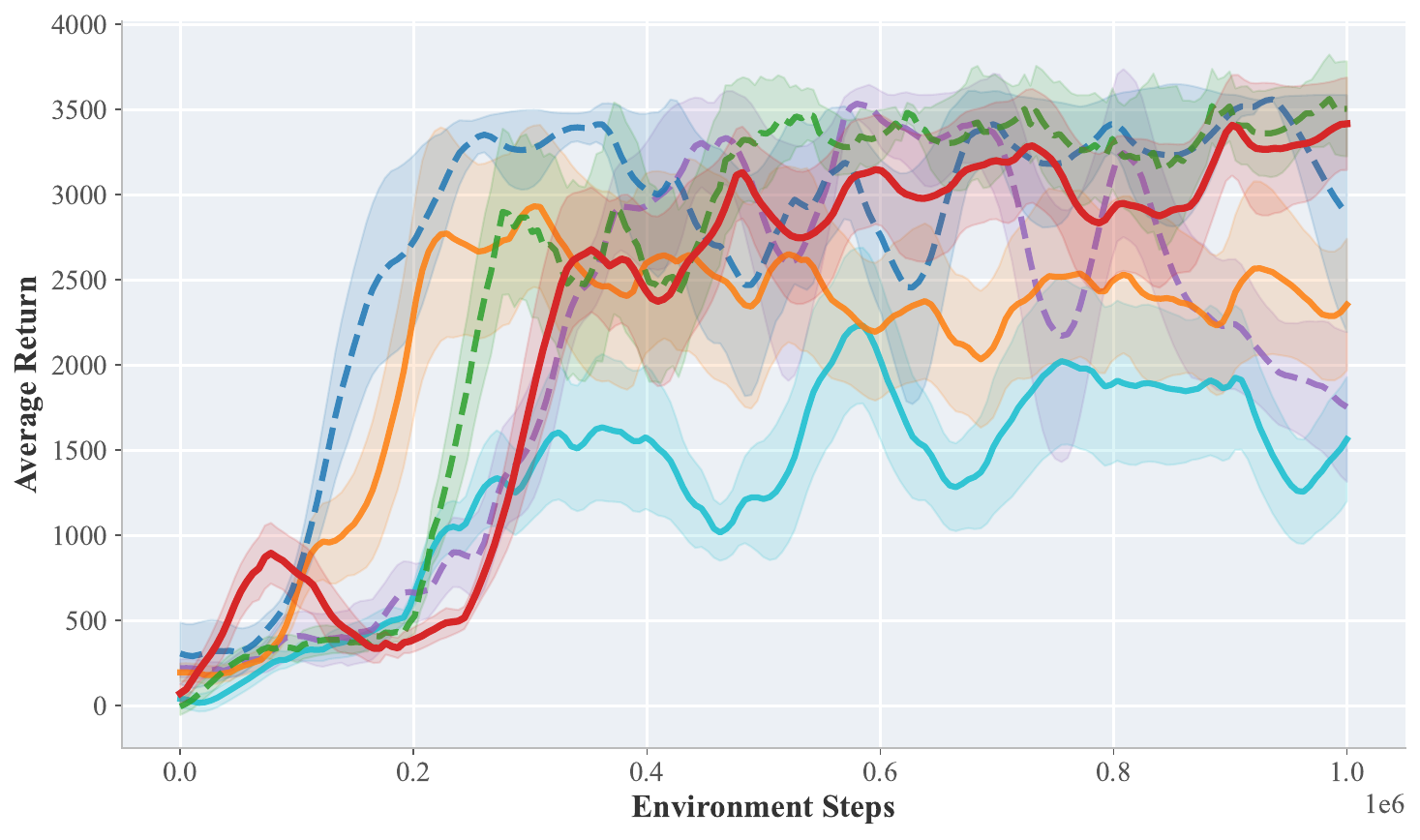}} \quad
\vspace{0.5em}
\subfloat[Humanoid-v4\label{subFig:humanoid}]
{\includegraphics[width = 0.3\textwidth]{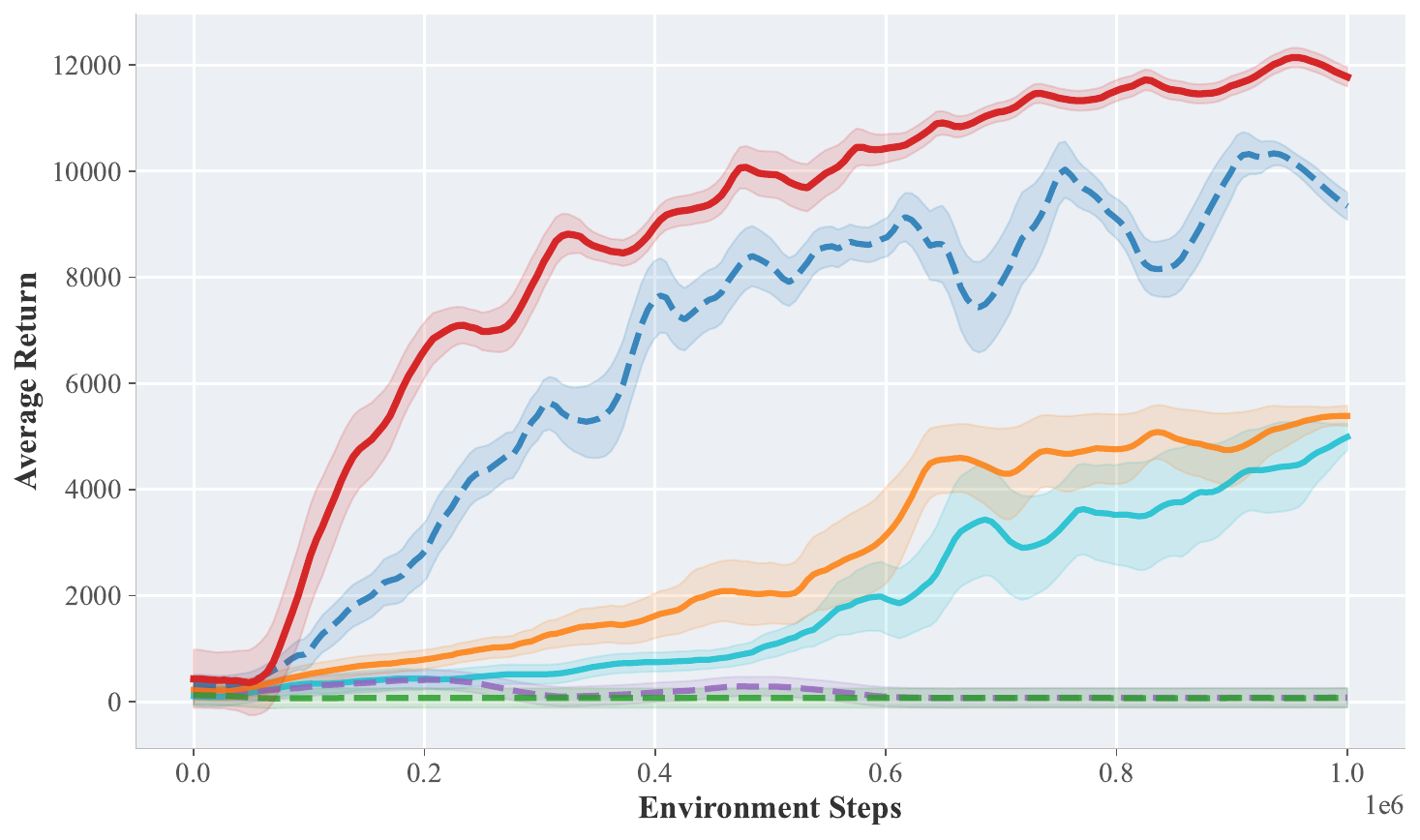}} \quad
\subfloat[InvertedPendulum-v4\label{subFig:ant}]
{\includegraphics[width = 0.3\textwidth]{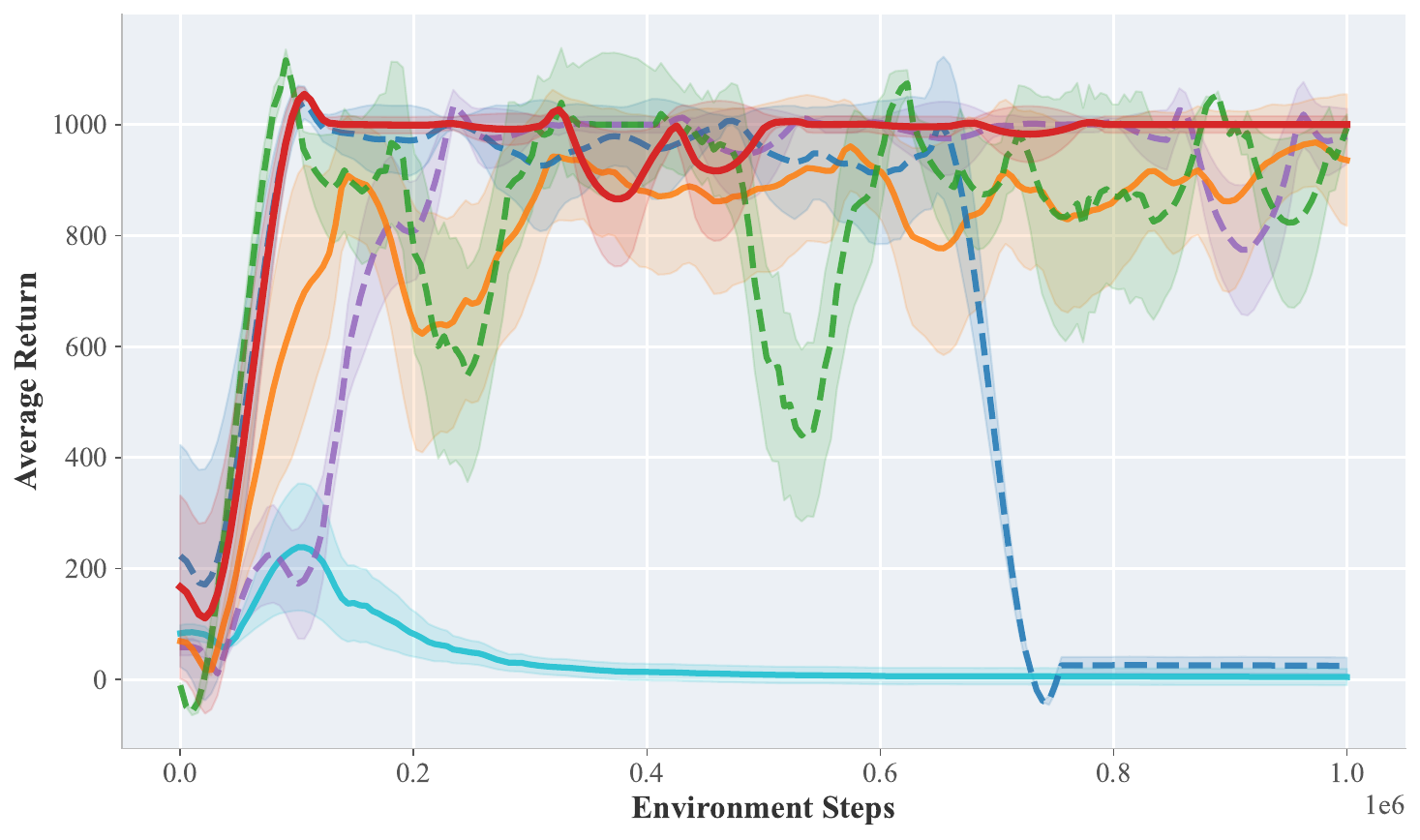}} \quad
\subfloat[Reacher-v4\label{subFig:Reacher}]
{\includegraphics[width = 0.3\textwidth]{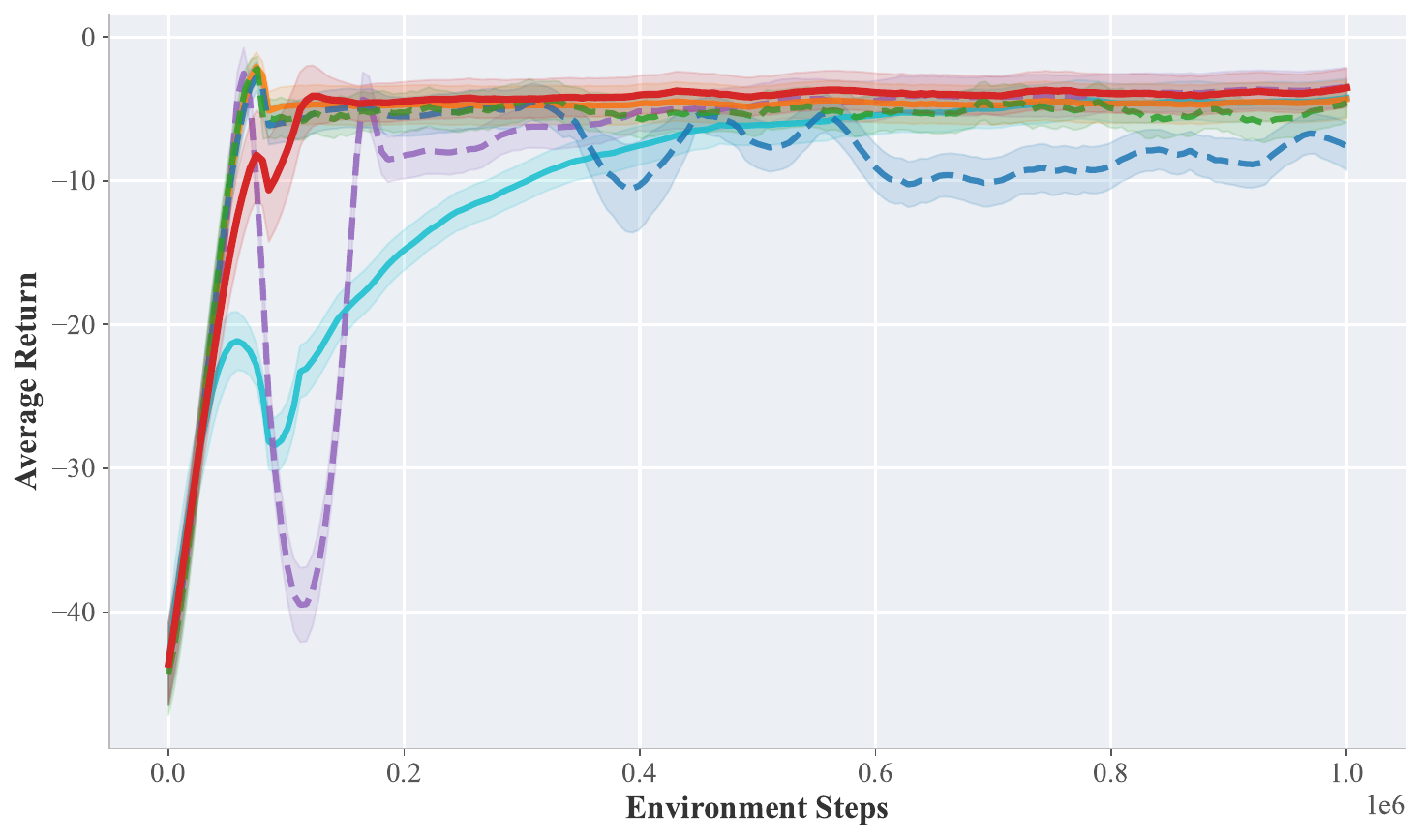}} \quad
{\includegraphics[width = 0.9\textwidth]{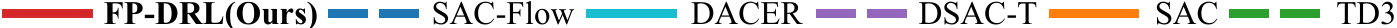}} \quad
\caption{Training curves on benchmarks. The solid lines correspond to mean and shaded regions
correspond to the standard error of the mean (SEM) over Three runs.}
\label{f:baseline}
\end{figure*}

We conducted three independent training runs for each experiment, each using a different random seed, with these random seeds kept consistent across all algorithms and benchmark tasks. The learning curves and policy performance are presented in Fig.~\ref{f:baseline} and Table~\ref{tab:fp-drl-comparison}, respectively. Our results demonstrate that FP-DRL outperforms, or at least matches, all baseline algorithms across all benchmark tasks. Taking Ant-v4 as an example, compared to SAC-Flow \cite{zhang2025sac}, DACER \cite{wang2024diffusion}, DSAC-T \cite{duan2025distributional}, SAC \cite{SAC}, and TD3 \cite{TD3}, our algorithm achieves relative performance improvements of 21.6\%, 81.6\%, 49.3\%, 20.2\%, and 72.6\%, respectively. These results establish FP-DRL as a new performance benchmark for model-free RL algorithms. Furthermore, compared to other flow-based methods such as SAC-Flow, significant improvements are achieved in both learning stability and final performance.

\begin{table*}[t]
\centering
\caption{Comparison of FP-DRL with baseline algorithms on MuJoCo continuous control tasks. Each value reports the mean $\pm$ standard deviation of episodic return over the last 10\% of training steps. The best value for each task is bolded.}
\label{tab:fp-drl-comparison}
\begin{tabular}{l c c c c c c}
\toprule
Task & FP-DRL & DSAC-T & SAC & TD3 & SAC-FLOW & DACER \\
\midrule
Ant & 
\textbf{6665} $\pm$ \textbf{0} & 
3499 $\pm$ 1081 & 
5179 $\pm$ 1408 & 
3023 $\pm$ 0 & 
5312 $\pm$ 1622 & 
5971 $\pm$ 410 \\
Half. & 
\textbf{15227} $\pm$ \textbf{512} & 
11694 $\pm$ 171 & 
13629 $\pm$ 994 & 
8824 $\pm$ 0 & 
12609 $\pm$ 0 & 
10277 $\pm$ 399 \\
Hopper. & 
3298 $\pm$ 608 & 
1958 $\pm$ 599 & 
2448 $\pm$ 769 & 
3319 $\pm$ 0 & 
\textbf{3338} $\pm$ \textbf{367} & 
1534 $\pm$ 619 \\
Human. & 
\textbf{11940} $\pm$ \textbf{0} & 
77 $\pm$ 2 & 
5184 $\pm$ 572 & 
69 $\pm$ 0 & 
10017 $\pm$ 0 & 
4553 $\pm$ 1134 \\
Invert. & 
\textbf{1000} $\pm$ \textbf{5} & 
860 $\pm$ 174 & 
941 $\pm$ 164 & 
778 $\pm$ 0 & 
25 $\pm$ 5 & 
5 $\pm$ 2 \\
Reacher & 
\textbf{-4} $\pm$ 0 & 
-4 $\pm$ \textbf{1} & 
-5 $\pm$ 1 & 
-5 $\pm$ 0 & 
-8 $\pm$ 2 & 
-5 $\pm$ 1 \\
\bottomrule
\end{tabular}
\end{table*}

\subsection{Ablation Study}
Subsequently, we carry out ablation studies to evaluate the impact of individual refinement within FP-DRL. In this section, we analyze why FP-DRL outperforms all other baseline algorithms on MuJoCo tasks. We conduct ablation experiments to investigate the impact of the following four aspects on the performance of FP-DRL: 1) whether to introduce a distributional critic to model the return as a distribution; 2) introducing a distributional critic but modeling the action output as a Gaussian distribution instead of using a flow-based policy; 3) the impact of different Transformer sequence lengths on performance; 4) the impact of different numbers of quantiles on performance. 
\begin{figure*}[htbp]
\centering
\captionsetup[subfigure]{justification=centering}
\subfloat[Only Distributional Critic\label{subFig:Only Flow-based Policy}]
{\includegraphics[width = 0.4\textwidth]{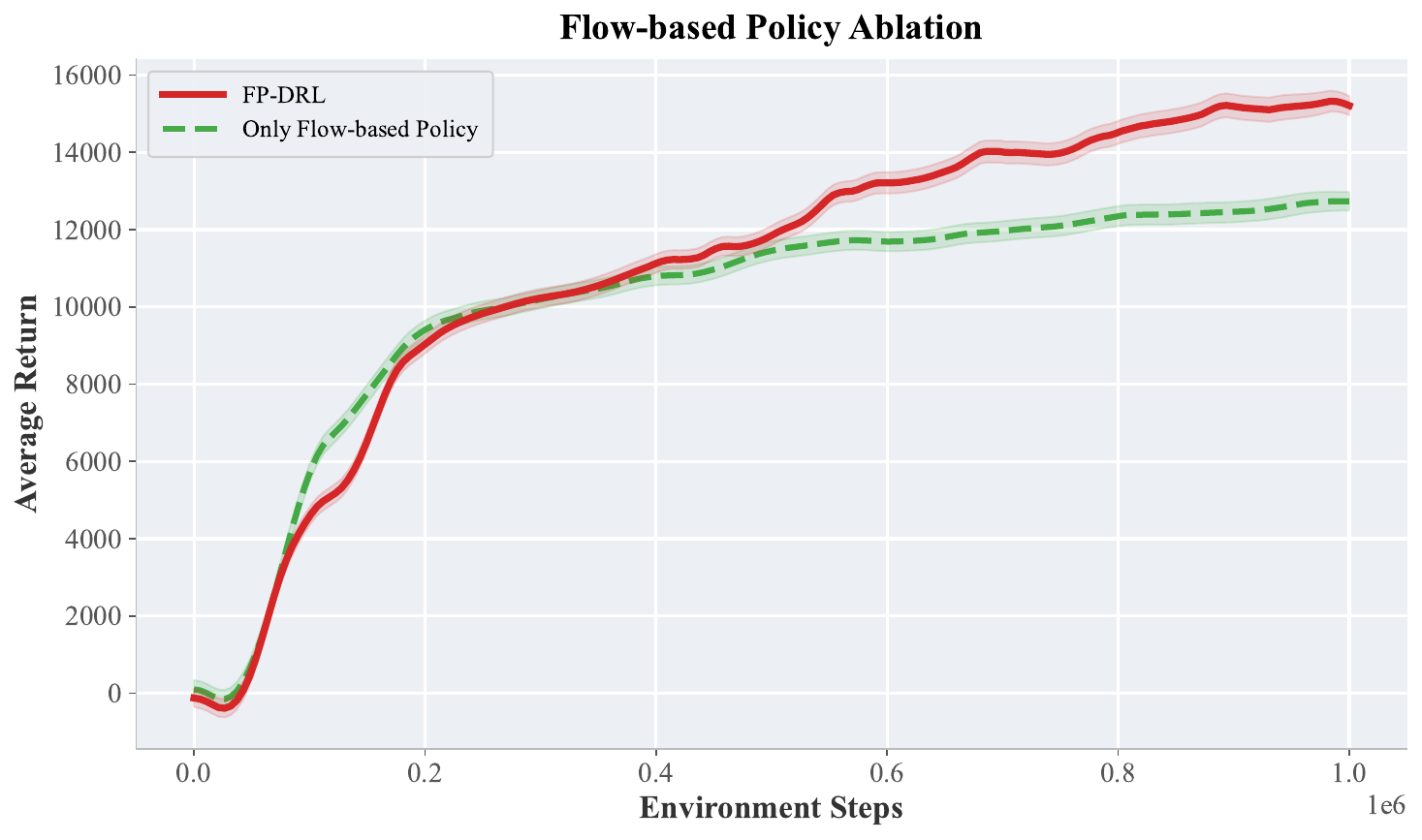}} \quad
\subfloat[Only Flow-based Policy\label{subFig:Only Distributional Critic}]
{\includegraphics[width = 0.4\textwidth]{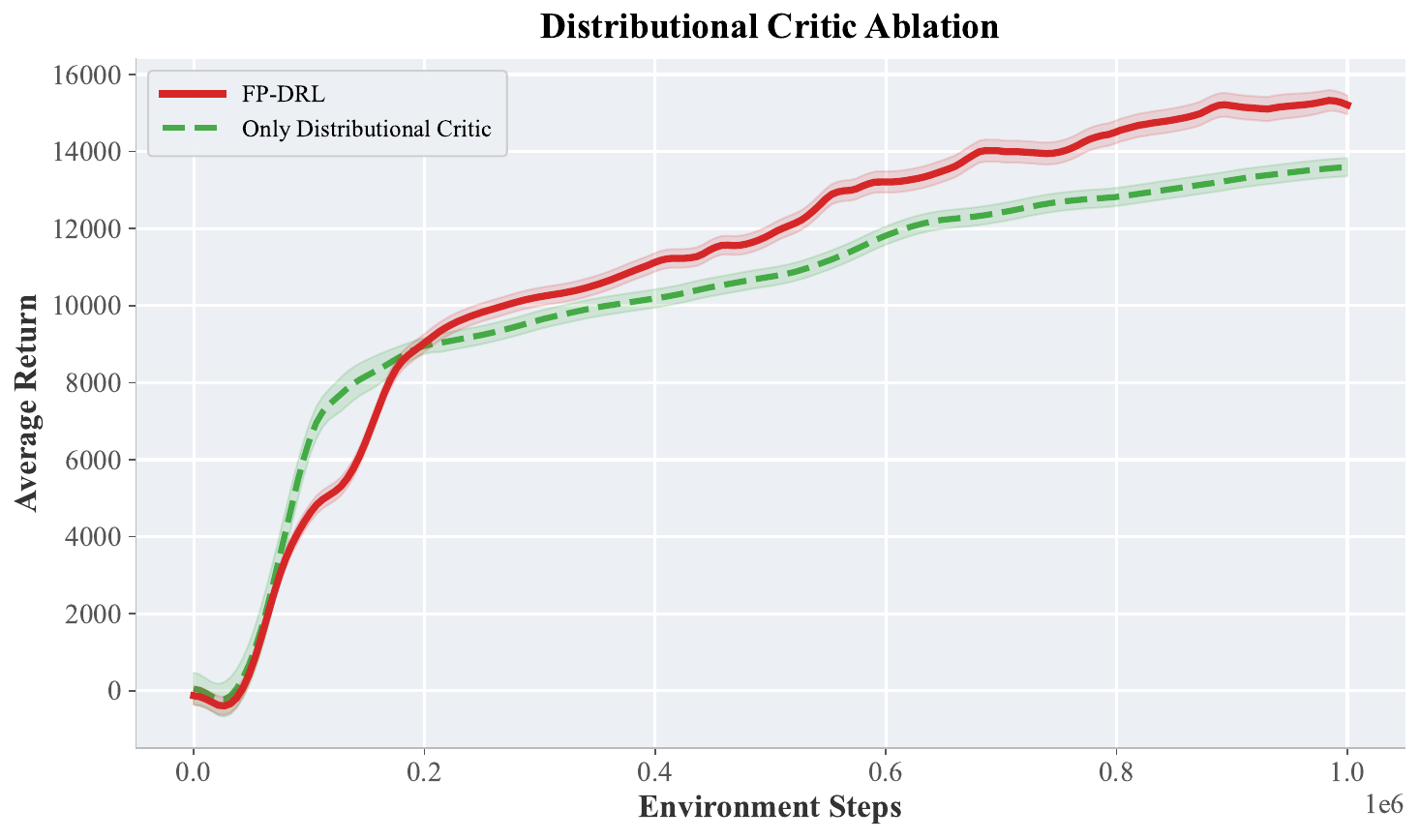}} \quad
\vspace{0.5em}
\subfloat[Quantiles number\label{subFig:Sequence Length}]
{\includegraphics[width = 0.4\textwidth]{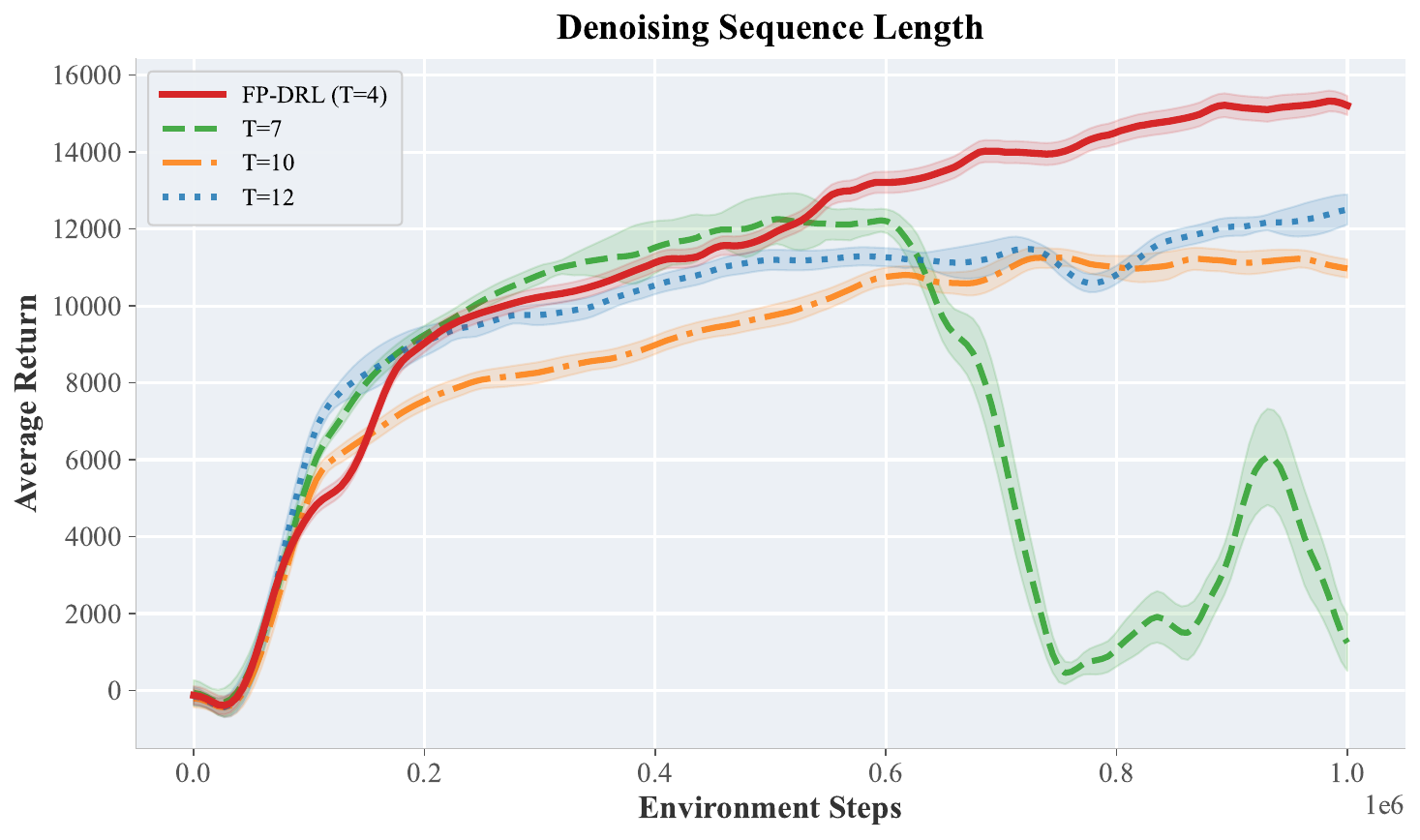}} \quad
\subfloat[Sequence Length\label{subFig:Quantiles number}]
{\includegraphics[width = 0.4\textwidth]{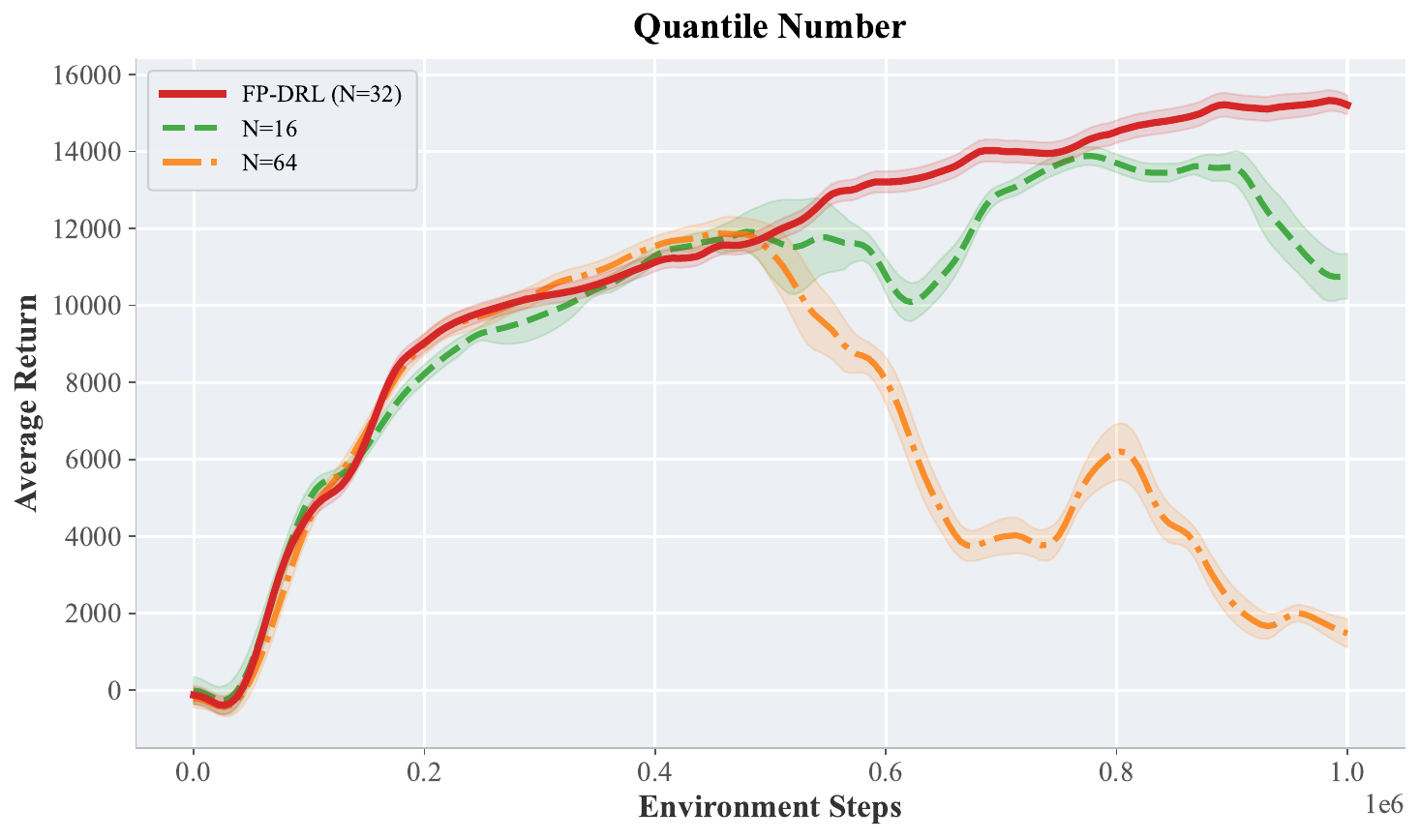}} \quad
\caption{Ablation studies of FP-DRL. (a) Performance comparison between a Gaussian policy and the proposed transformer-based flow policy. (b) Comparison of modeling returns using the mean versus a quantile-based distributional critic. (c) Effect of varying the number of quantiles (N=16, 32, and 64) for the distributional critic. (d) Training curves across different transformer sequence lengths (K=4, 7, 10, and 12).}
\label{f:ablation}
\end{figure*}

\noindent\textbf{Only Flow-based Policy.} To validate the significant role of introducing a distributional critic in final performance, we conducted the following two experiments on the HalfCheetah-v4 task: 1) modeling returns using the mean; 2) modeling the return distribution using a quantile-based approach. As shown in Fig.~\ref{subFig:Only Flow-based Policy}, modeling the return distribution with a quantile-based approach provides more effective guidance for the flow-based policy, achieving the best performance.

\noindent\textbf{Only Distributional Critic.} To validate the significant role of the flow matching-based policy in performance, we conducted two experiments under the distributional critic setting: 1) modeling the policy using a Gaussian distribution; 2) modeling the policy using a transformer-based flow model. As shown in Fig.~\ref{subFig:Only Distributional Critic}, Flow-based Policy with distributional critic achieves the best performance.

\noindent\textbf{Sequence Length.} We further investigated the performance of the flow-based policy under varying transformer sequence lengths. On the HalfCheetah-v4 task, we plotted training curves for sequence lengths of K=4, 7, 10, and 12. As shown in Fig.~\ref{subFig:Sequence Length}, sequence length of 4 achieves the best performance. Thus, we selected a sequence length of 4 for all experiments. 

\noindent\textbf{Quantiles number.} We also examined the effect of the number of quantiles on the performance of the distributional critic. On the HalfCheetah-v4 task, we plotted training curves for N=16, 32, and 64. As shown in Fig.~\ref{subFig:Quantiles number}, Quantiles number of 32 achieves the best performance. Thus, we selected a quantiles number of 32 for all experiments.

\section{Conclusion}
\label{sec:conclusion}
In this paper, we introduced FP-DRL, a novel RL framework that synergizes the expressive power of Flow Matching with the robust guidance of distributional RL. By parameterizing the policy as a flow-based model, we overcome the inherent limitations of traditional Gaussian policies, enabling the agent to capture complex, multimodal action distributions with high computational efficiency. Furthermore, by shifting from scalar expected values to a quantile-based return distribution, our approach provides the policy with a richer, more discriminative signal that accounts for the aleatoric uncertainty and multimodality of rewards.

\begin{credits}
\subsubsection{\ackname}This work is supported by National Natural Science Foundation of China (No.
72301289) and National Natural Science Foundation of China (No.62306335). We thank all the anonymous reviewers who generously contributed their time and efforts. Their professional recommendations have greatly enhanced the quality of the manuscript.

\subsubsection{\discintname}
The authors declare that they have no known competing financial interests or personal relationships that could have appeared to influence the work reported in this paper.
\end{credits}
\bibliographystyle{splncs04}
\bibliography{ref}
%





% \end{thebibliography}
\end{document}